\pdfoutput=1

\documentclass[11pt]{article}

\usepackage{acl}

\usepackage{times}
\usepackage{latexsym}

\usepackage[T1]{fontenc}

\usepackage[utf8]{inputenc}

\usepackage{microtype}

\usepackage{times}
\usepackage{latexsym}
\usepackage{comment}
\usepackage{graphicx}
\usepackage{subcaption}
\usepackage{amssymb}
\usepackage{amsfonts}
\usepackage{url}
\usepackage{pifont}
\usepackage{multirow}
\usepackage{arydshln}
\usepackage[normalem]{ulem}

\newcommand{\old}[1]{}

\usepackage{xspace}
\newcommand{\ArgSciChat}{ArgSciChat\xspace}
\newcommand{\e}{\texttt{E}\xspace}
\newcommand{\p}{\texttt{P}\xspace}

\usepackage{booktabs}
\usepackage{cleveref}

\title{ArgSciChat: A Dataset of Argumentative Dialogues on Scientific Papers}

\author{Federico Ruggeri$^1$ \thanks{\hspace{2mm}Work done while conducting an internship at UKP.} \\
  \And
  Mohsen Mesgar$^2$ 
  \\
    $^1$DISI, University of Bologna \\
  $^2$Ubiquitous Knowledge Processing Lab (UKP Lab) \\
  Department of Computer Science \\
  Technical University of Darmstadt \\
  $^1$\texttt{federico.ruggeri6@unibo.it} \\
  $^2$ \texttt{www.ukp.tu-darmstadt.de} \\
  \And Iryna Gurevych$^2$
  }

\begin{document}
\maketitle

\begin{abstract}
With recent advances in question-answering models, various datasets have been collected to improve and study the effectiveness of these models on scientific texts. 
Questions and answers in these datasets aim to explore the content of a scientific paper by seeking information from the paper's content. 
However, these datasets do not tackle the argumentative content of scientific papers, which is of huge importance in persuasiveness of a scientific discussion. 
We introduce ArgSciChat, a dataset of 41 argumentative dialogues between scientists on 20 NLP papers. 
The unique property of our dataset is that it includes both exploratory and argumentative interactions in a dialogue on a scientific paper.  
Moreover, the size of ArgSciChat demonstrates the challenges in collecting dialogue for specialized domains. 
Thus, our dataset is an aid for evaluating few-shot transfer learning methods for dialogue agents in a specialized domain. 
We perform a detailed analysis on \ArgSciChat and use it to fine-tune and evaluate a pre-trained document-grounded dialogue agent for NLP papers.  
The results show that dialogues in ArgSciChat include exploratory and argumentative interactions, which makes it challenging for the examined dialogue agent to answer questions. 
We publicly release \ArgSciChat.
\footnote{\url{https://github.com/UKPLab/arxiv2022-argscichat}} 

\end{abstract}

\begin{figure*}[!t]
    \centering
    \includegraphics[width=1.0\linewidth]{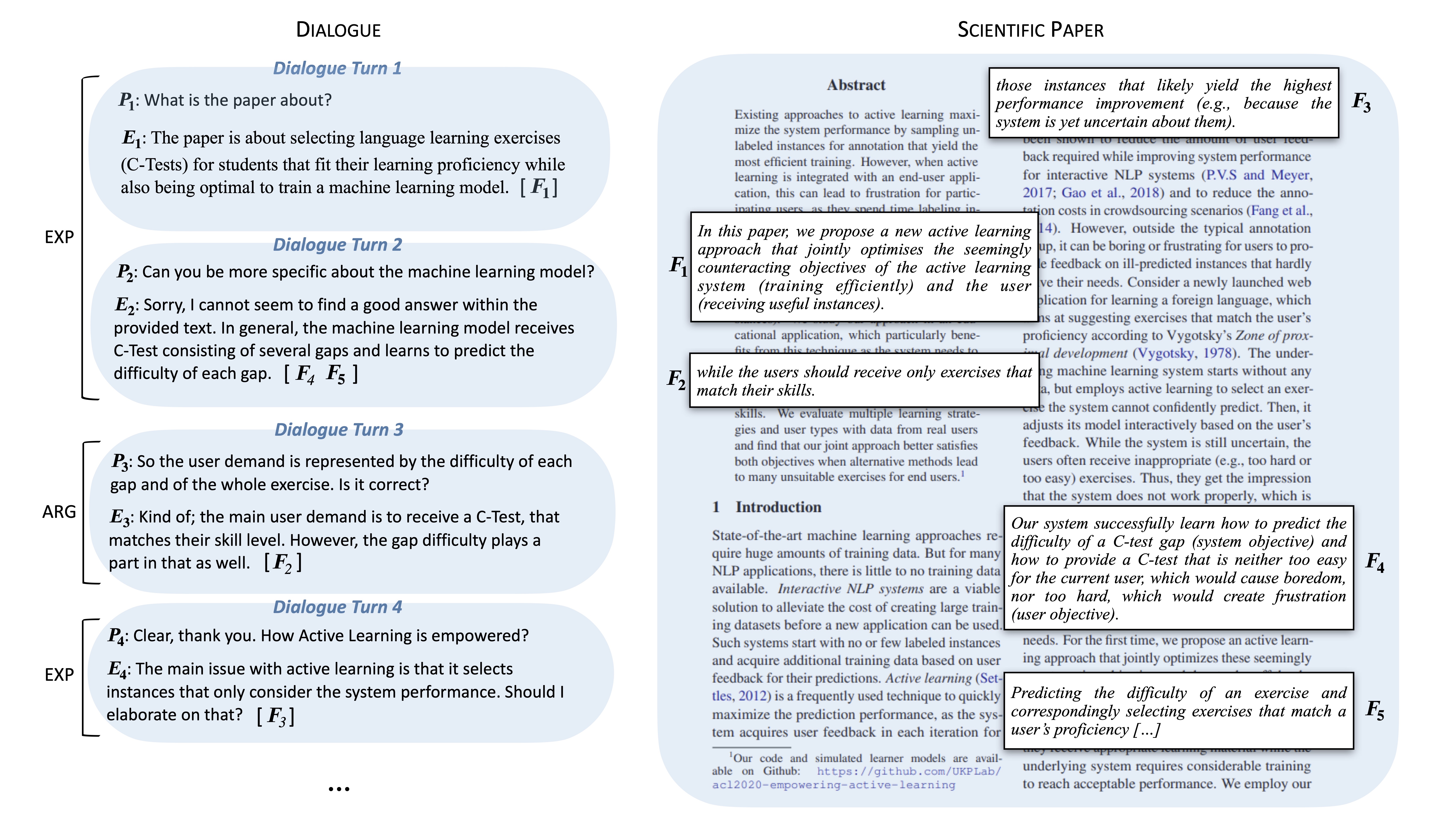}
    \caption{An example dialogue from our \ArgSciChat\ dataset. 
        Supportive facts $\{F_1, \dots, F_6\}$ are highlighted in the scientific paper. 
        Dialogue partners alternate between exploratory (EXP) and argumentative (ARG) intents. 
    }
    \label{fig:conversation_example}
\end{figure*}

\section{Introduction}
\label{sec:introduction}

The enormous and ever-growing number of scientific papers~{\cite{Munroe2013TheRO, ronzano15}} make scientific text processing imperative. 
A considerable body of research in NLP is dedicated to developing methods to provide insights from scientific papers~\cite{beltagy-etal-2019-scibert,zhang-etal-2021-abstract,wadden-etal-2022-multivers,zhang-etal-2021-abstract,parveen-etal-2016-generating,wadden-etal-2022-multivers,mysore-etal-2022-multi}.
%
%
Recent advances in Question-Answering (QA) models made them a great asset in accessing the content of scientific papers~\cite{Dasigi2021ADO}. 
To develop and evaluate QA models on scientific papers, various datasets have been collected~\cite{Dasigi2021ADO,DBLP:journals/corr/abs-1809-00732,pappas-etal-2018-bioread,DBLP:journals/corr/abs-1909-06146,DBLP:journals/bmcbi/TsatsaronisBMPZ15, 10.1007/978-3-030-45442-5_71}. 
For each scientific paper in these datasets, a set of questions with the exploratory (EXP) intention is defined. 
These questions aim to explore the content of a scientific paper and seek information. 
Answers to EXP questions are limited to single words or short text spans extracted from the text of the paper. 

Scientific answers should ideally be explained with argumentation to be persuasive. 
If an answer is not persuasive, then a follow-up question might be raised to argue about the answer. 
This property of scientific domain motivates a crucial need for a dataset of scientific argumentative dialogues on scientific papers. 
Collecting such a dataset is challenging for two reasons.
First, while there are linguistic theories about scientific argumentative dialogues~\cite{walton_2008,macagno2000types}; there is no functional intent sets for collecting argumentative  questions and answers.  
Second, scientists (and experts in general) are on a strictly limited time-budget, so incentivizing scientists and aligning them to chat about a scientific paper should be very optimized. 
Although off-the-shelf crowdsourcing platforms are applicable for data collection in generic domains, they do not lend themselves for collecting argumentative dialogues in science. 

In this work, we introduce \emph{\ArgSciChat}, the first dataset of annotated argumentative dialogues on NLP papers. 
Figure~{\ref{fig:conversation_example}} shows an example dialogue from our dataset. 
To collect ArgSciChat, we first define a set of intent classes for interactions between scientists about scientific papers. 
We then develop a new dialogue collection methodology in which we involve scientists preferences into the dialogue collection process. 
We allow scientists to choose a paper to chat about based on their preference. 
We then align them with each other to conduct a dialogue about their chosen papers according to their schedule. 
In ArgSciChat, the dialogues are in English and conducted synchronously by \emph{subjects} who are NLP scientists. 
Each dialogue is about a single paper. 
We define the aim of dialogues to exchange information and argue about retrieved answers from the content of a paper. 
Each subject plays a different role in the dialogue.  
\e is the \textbf{expert} on the research presented in the paper, and \p (for \textbf{proponent}) is the subject aiming to learn about the scientific paper by posing questions to the expert. 
Importantly, \e and \p can argue about the content of the paper and each other's opinion. 

We extensively analyze the properties of the collected dialogues in ArgSciChat.   
To put our dataset in context, we compare our dataset with CoQA \cite{reddy2019coqa}, QuAC~\cite{choi-etal-2018-quac} and Doc2Dial~\cite{DBLP:journals/corr/abs-2011-06623}.   
We then evaluate the diversity of dialogues in our dataset and analyze the quality of answers. 
Finally, we use our dataset to fine-tune and evaluate LED \cite{Beltagy2020LongformerTL}, a generic dialogue agent for answering questions about a long text\old{ using Longformers}. 
The goal of this experiment is to study the difficulty level of our dataset for current dialogue systems. 

In summary, our major contributions are 
\textbf{(1)} a high-quality and challenging argumentative dialogue dataset on NLP papers, 
\textbf{(2)} a new set of intent classes for interactions in argumentative dialogues, 
\textbf{(3)} a new methodology for collecting dialogues between scientists.

We believe that our contributions pave the way for building intelligent dialogue agents in scientific domains. 
These agents not only can foster our access to the content of scientific papers by answering our questions but also can scientifically argue about their answers. 

\begin{table*}[!tb]
\small
\centering
\begin{tabular}{@{}p{2.4cm}p{5.5cm}p{7.5cm}@{}}
\toprule
\textbf{Intent} & \textbf{Definition} & \textbf{Example} \\
\midrule
\multicolumn{3}{l}{\textbf{Exploratory (EXP)}} \\
Ask Info (AI)      & A question to seek information from a  paper. & \emph{What data set is used to train a classifier?} \\
Give Info (GI)    & An answer to give information about a paper. & \emph{We consider the Partial Latin Square completion problem.}  \\
Switch Topic (ST) & A request or proposal to shift the topic. & \emph{Which intent would you like to know more about?} \\
\midrule
\multicolumn{3}{l}{\textbf{Argumentative (ARG)}} \\
Ask Opinion (AO)  & A question to ask about opinions.           & \emph{Which do you think is the strongest point of your approach?} \\
Give Opinion (GO)  & An answer to provide an opinion.    & \emph{It is interesting that such simple techniques help.} \\
\bottomrule
\end{tabular}
     \caption{We define five intent classes for questions and answers in a scientific argumentative dialogue.}
    \label{tab:action_set}
\end{table*}

\section{Related Work}
\label{sec:related_work}

To our knowledge, there is no dataset consisting of dialogues with both exploratory (EXP) and argumentative (ARG) interactions.  
The most similar datasets to ours deal with QA tasks on scientific papers, such as QASPER~\cite{Dasigi2021ADO}, emrQA \cite{DBLP:journals/corr/abs-1809-00732}, BioRead \cite{pappas-etal-2018-bioread}, BioMRC \cite{DBLP:journals/corr/abs-2005-06376} and QAngaroo \cite{welbl-etal-2018-constructing}, PubMedQA \cite{DBLP:journals/corr/abs-1909-06146}, and BioAsq \cite{DBLP:journals/bmcbi/TsatsaronisBMPZ15, 10.1007/978-3-030-45442-5_71}. 
These datasets include only the EXP questions and lack the ARG questions and answers. 
Moreover, questions in these datasets weakly link to each other to resemble a dialogue.  
In contrast, \ArgSciChat\ consists of coherent dialogues with both EXP and ARG questions. 
Answers in the aforementioned datasets are limited to single words or text spans extracted from a paper's content. 
Conversely, dialogues in \ArgSciChat are exclusively in a free-format text, a major property of human-like dialogue~\cite{reddy2019coqa}. 

In the literature, there are a few datasets that contain dialogues grounded on non-scientific documents, e.g., Wikipedia, conducted by non-expert crowd subjects. 
In particular, we refer to CoQA~\cite{reddy2019coqa}, QuAC~\cite{choi-etal-2018-quac}, and Doc2Dial~\cite{DBLP:journals/corr/abs-2011-06623}. 
We take inspiration from their methodology to define our own for collecting dialogues between scientists.
The advantage of our method is that it overcomes the challenge of collecting synchronous dialogues between scientists given their limited time budget. 
Our methodology allows scientists to choose a paper based on their scientific preference and participate in a dialogue based on their time budget.  
More importantly, unlike these datasets, dialogues in \ArgSciChat take into account both EXP and ARG interactions.

The theory of argumentative dialogues was addressed many years ago~\cite{macagno2000types}. 
However, argumentative datasets concerning a conversational setting have been scarcely explored~\cite{lawrence2020}. 
Existing argumentative datasets are built for
argument classification~\cite{chakrabarty-etal-2019-ampersand,al-khatib-etal-2016-cross,DBLP:conf/acl/HaddadanCV19, park18, orbach2019}, extraction~{\cite{visser2020, janierR16}}, ranking~\cite{habernal-gurevych-2016-argument,lauscher2020rhetoric}, summarization~{\cite{roush2020}}, and generation~\cite{hua-wang-2018-neural, hua2019sentencelevel}. 
Unlike these datasets, our dataset includes scientific argumentative dialogues where the arguments should be inferred from the content of scientific papers.

\section{Methodology}
\label{sec:method}
Our goal is to collect synchronous dialogues about the content of scientific papers, where dialogues  include EXP and ARG questions and answers.

\subsection{Defined Intent Set} 
\label{sec:dialogue_collection}
There are no pragmatical intent sets for collecting scientific argumentative dialogues. 
Therefore, we define a new set of intents for interactions in such dialogues. 
To do so, we follow a fundamental theory of argumentative dialogues~\cite{walton_2008}. 
\newcite{walton_2008} defines six different interaction types in a dialogue. 
By taking into account their goals, only two types are relevant to scientific dialogues: information-seeking to extract information and persuasion to discuss the claims of a scientific paper. 
Inspired by this theory, we define five intent classes for questions and answers that may happen in scientific argumentative dialogue. Table~{\ref{tab:action_set}} summarizes the definitions of these intents and shows an example for each intent.
Exploratory (EXP) interactions aim to seek (or provide) information. 
Argumentative (ARG) interactions aim to convey persuasion.
It is worth noting that \newcite{walton_2008} considers other aspects of argumentative dialogues, such as ``personal attacks'' and ``appeals to emotion''. 
We do not include those aspects in this work because scientific argumentation is primarily supposed to be objective.   

\subsection{Dialogue Formulation} 
\label{sec:data_collection_pipeline}
In each dialogue, each subject receives either role \e (expert) or \p (proponent). 
\p and \e discuss the content of a scientific paper. 
\p accesses only the paper's title, whereas \e accesses the abstract and introduction sections of the scientific paper.
We limit the paper content to the abstract and introduction sections because we believe these two sections present the gist of the paper sufficiently to sustain a dialogue.

To collect EXP interactions, we instruct \p to use the title to start the conversation by asking for information about the paper. 
\e answers the question using the content of given sections of the paper. 
Since \e's answers should be grounded in the content of the scientific paper, we ask \e to select up to two text spans from the paper that are used in generating the answer.  
We refer to these text spans as supporting facts. 
This definition resembles the ones introduced in {\citet{Dasigi2021ADO}} (evidence) and {\citet{reddy2019coqa}} (rationale). 
We emphasize that a supporting fact does not imply any argumentative function of the text span (e.g., a fact can be a claim like $F_5$ in Figure~{\ref{fig:conversation_example}}). 
A supporting fact only helps the subjects to conduct high-quality dialogues. 
To encourage natural conversations, we instruct both \p and \e to write down their message (which could be a question or an answer) in free-form textual sentences. 

A pair of a \p's and an \e's messages constitutes a dialogue turn. 
The \e's message and selected supportive facts are displayed to \p to initiate the next dialogue turn. 
To collect messages with ARG intent, we encourage \p and \e to argue about previous dialogue turns. 
An explored topic of the scientific paper should be easier to understand through argumentation. 
Finally, to limit the occurrences of unanswerable questions and improve the coherence of the dialogues, we ask \e to switch the topic of conversation using the given content of the paper.

\subsection{Dialogue Collection}
\label{sec:implementation}
To collect scientific argumentative dialogues on scientific papers, we focus on NLP publications. 
While including many interdisciplinary scientific papers, this domain sets a common background between scientists as subjects.
For each dialogue we ask two subjects who are experts in NLP to discuss about the content of an NLP publication. 

A key challenge in implementing our data collection methodology is to optimize this process for scientists. 
Best-practice crowdsourcing frameworks (e.g., AMT and Upwork) lack three main features regarding dialogue collection in the scientific domain: (1) flexibility in scientific paper selection and (2) time slots scheduling, and (3) synchronous participation. 
To meet these requirements and incentivize scientists' participation, we introduce an implementation for our dialogue collection methodology. 
We develop a web-based tool where the subjects can register and chat in a written format about an NLP publication. 
While subjects sign up in our tool, they confirm a consent which authorizes the usage of their dialogues and papers for any sorts of research purposes. 
We define the consent form by following GDPR\footnote{\url{https://gdpr.eu/tag/gdpr/}} and the ethics guidelines for trustworthy AI\footnote{\url{https://op.europa.eu/en/publication-detail/-/publication/d3988569-0434-11ea-8c1f-01aa75ed71a1}}. 

To encourage subjects participate in a dialogue, we let them select a few scientific papers from an automatically retrieved list of their recent publications. 
This idea relieves subjects' burden in reading the selected paper since they are the author of those papers. 
Moreover, from the subjects' perspective, participating in a dialogue about their publications is an informal advertisement for their research. 
To align the subjects with each other time-wise, we ask subjects with role \e to announce their available time slots for presenting their selected scientific paper. 
We then let subjects with role \p choose time slots based on their time schedule and also the title of papers associated to the slots. 
We report the implementation details of our methodology and an example of its web interfaces in Appendix~\ref{appendix:collection_details}.

We invited 31 senior and junior scientists in NLP from two large NLP groups in Europe to participate in our dialogue collection study.   
23 of the invited scientists (74.2\%) accepted and participated in at least one dialogue. 
On 20 NLP papers, we collected 68 dialogues in total.  
We filtered out dialogues that ended abruptly due to connection issues, resulting in 41 dialogues.
We present these dialogues and their corresponding papers as the \ArgSciChat dataset, which consists of 498 messages from 41 dialogues about 20 NLP papers. 
\section{Experiments} 
\label{sec:dialogues_analysis}
\subsection{Comparing with Similar Datasets}
To put the \ArgSciChat dataset in context, we first compare it with similar datasets.      
We compare dialogues in \ArgSciChat with dialogues in 
CoQA~\cite{reddy2019coqa}, QuAC~\cite{choi-etal-2018-quac}, and Doc2Dial~\cite{DBLP:journals/corr/abs-2011-06623}. 
These datasets include contextualized question-answering (QA) messages grounded in generic documents.
Since no expert knowledge is required to explore such texts, dialogues take place between crowd contributors in these datasets. Conversely, dialogues in \ArgSciChat are conducted by experts in NLP as a scientific domain. Importantly, unlike the other datasets, dialogues in \ArgSciChat follow EXP and ARG intents. 

\subsection{Analyzing Dialogues of ArgSciChat}
To obtain a deep understating of the dialogues in ArgSciChat, we perform an in-depth analysis on this dataset.
In particular, we investigate the diversity of dialogues collected for a paper, the properties of selected supportive facts, and to what extent the dialogues are exploratory and argumentative. 

\subsection{Using ArgSciChat for Evaluating Dialogue Agents}
To assess the difficulty of \ArgSciChat for recent dialogue agents, we synthesize a dialogue agent to take the role of \e. 
Consequently, we define two evaluation tasks:
\textbf{(i) Fact selection}, where the agent selects up to two sentences from a given scientific paper as supportive facts; and 
\textbf{(ii) Response generation}, where the agent generates a free-form text to respond to a \p's message. 

We evaluate LED~\cite{Beltagy2020LongformerTL}, which is a Longformer trained to answer questions about the content of a long text.  
LED has been recently used for QA on scientific papers~\cite{Dasigi2021ADO}.
Using our dataset, the input to LED can be: a \p's question to which we refer as a query (Q);  an NLP paper (P); all interactions exchanged before Q, also known as dialogue history (H); 
and the supportive facts (F) selected by human subjects to generate the reference \e's answers. 
We thus compare the following LED configurations:
LED(Q, P), LED(Q, P, H), and LED(Q, F), where terms inside the parentheses are the inputs to the LED agent.  

\newcite{Dasigi2021ADO} show that LED achieves better performance when trained jointly for response generation and fact selection. 
Likewise, we train LED using the loss function $\mathcal{L} =  \mathcal{L}_{gen} + \mathcal{L}_{fact}$,
where $\mathcal{L}_{gen}$ is the average negative log-likelihood loss between a generated message and the reference \e message.
The $\mathcal{L}_{fact}$ loss is the average binary cross-entropy loss between the label that a neural classification layer, i.e., MLP, assigns to each sentence of the paper and the reference fact labels we defined for those sentences.
To define the reference fact labels for a sentence in a scientific paper,  we compare the sentence with the corresponding human-selected supportive facts.
We assign a ground truth label 1 to a sentence if a supportive fact contains that sentence, otherwise 0.

We carry out a five-fold cross-validation routine for this experiment. 
We create fold splits such that all dialogues about the same scientific paper are in the same split.
Table~\ref{tab:folds} shows the number of dialogue turns as data samples used in the training, validation, and test sets of each fold. We report the values of all hyper-parameters used to train LED in Appendix~\ref{appendix:experiments}. 
\begin{table}[!t]
\small
\centering
\begin{tabular}{@{}lccccc@{}}
\toprule
      & \textbf{Fold 1} & \textbf{Fold 2}    & \textbf{Fold 3}    & \textbf{Fold 4}    & \textbf{Fold 5}    \\ 
\midrule
Train      & 199 & 200 & 183 & 212  & 207 \\
Validation & 27 & 14  & 39 & 27 & 22 \\
Test       & 23 & 35 & 27 & 10 & 20 \\ 
\bottomrule
\end{tabular}
\caption{The number of dialogue turns in each fold.
}
\label{tab:folds}
\end{table}

For evaluation metrics, similar to \newcite{Dasigi2021ADO}, for fact selection, we compute the F1 score over candidate sentences in a scientific paper against the reference supportive facts.
We denote this metric as \textit{Fact-F1}. 
For response generation, we use the token-level F1 score introduced in SQUAD~{\cite{rajpurkar-etal-2016-squad}}. 
This metric is computed over individual words between the generated response and the reference \e's message. 
We denote this metric as \textit{Message-F1}.
Following recent suggestions for evaluating text generation systems, we also report BERTScore (\textit{BScore})~\cite{ZhangKWWA20} and MoverScore (\textit{Mover})~\cite{ZhaoPLGME19} for the response generation task. 

To put our results in context, we use a retrieval-based method as a baseline for the fact selection task. 
This method computes the cosine-similarity score between the vector representation of a query Q and the vector representation of each sentence in the input paper.  
Following \citet{Dasigi2021ADO}, we use TF-IDF with default parameters in \textit{sklearn} and S-BERT to obtain the vector representations. 
We then rank the sentences concerning their similarity scores with the query Q and select the top-two sentences as the supportive facts. 
Like~\citet{Dasigi2021ADO}, we estimate a statistically robust human baseline for the fact selection task with three NLP experts. 
In particular, each expert selects two sentences from the input paper as supportive facts for a given query Q.
We would like to remark that human performance estimation process is independent of the annotation process described in Section~\ref{paragraph:role-specific_behaviours} and involves different NLP experts to ensure sound evaluation.

\section{Results}
\label{sec:results}

\subsection{Comparing with Similar Datasets}
Table \ref{tab:dialogue_analysis} reports the characteristics of dialogues in the compared datasets.
In terms of the average number of dialogue turns, dialogues in \ArgSciChat are comparable with those in QuAC and Doc2Dial datasets.
Compared with CoQA, dialogues in \ArgSciChat contain less dialogue turns. \ArgSciChat contains fewer dialogues than the other datasets. 
The limited number of dialogues in our dataset is symptomatic of expert domains where the data collection process is subject to tight requirements like matching scientists' schedules.
The exchanged messages in \ArgSciChat are beyond text spans and consist of free-form textual sentences.  
In particular, \ArgSciChat contains a higher percentage (50\%) of multi-sentence messages than the other datasets.
This observation is also confirmed by the average message length, i.e., the number of tokens per message. 

\begin{table}[!t]
\small
\centering
\begin{tabular}{lcccc}
\toprule
\textbf{Dataset}    & \begin{tabular}[c]{@{}c@{}}\textbf{Avg.}\\ \textbf{Turns}\end{tabular} 
& \textbf{\# Dial.} 
& \textbf{\% MSM} 
& \begin{tabular}[c]{@{}c@{}}\textbf{Avg.} \\ \textbf{Length}\end{tabular}  \\ \midrule
CoQA       & 15.5                        & 8k                         & 0.2\%  & 4.7                                                           \\
QuAC       & 7.3                         & 13.5k                         & 4.0\%  & 11.4                                                       \\
Doc2Dial   & 6.4                         & 4.8k                          & 17.8\% & 16.3                                                       \\
ArgSciChat & 6.3                         & 41                            & 50.8\% & 38.2                                                       \\
\bottomrule
\end{tabular}
\caption{
\textbf{Avg. Turns} shows the average number of turns per dialogue. 
\textbf{\% MSM} depicts the percentage of multi-sentence message in a dataset. 
\textbf{Avg. Length} indicates the average number of tokens per message \old{in a dataset}. \textbf{\# Dial.} indicates the number of dialogues. Supporting facts are not considered as answers.
}
\label{tab:dialogue_analysis}
\end{table}

\subsection{Analyzing Dialogues of ArgSciChat}
\label{sec:dialogue-diversity}
\paragraph{Dialogue Diversity.} 
On average \ArgSciChat contains two dialogues for each NLP paper. 
We study to what extent the messages (which could be questions or answers) exchanged in these dialogues are semantically diverse. 
To do so, we group messages in dialogues grounded in one paper into three categories:  
\textbf{G1:} semantically similar \p messages; 
\textbf{G2:} \e messages associated with \p messages in G1;  
and
\textbf{G3:} \e messages and their corresponding supportive facts. 
For each group, we compute the semantic diversity between any pair of messages, using the cosine distance between the vector representations of messages. 
We obtain the vector representations of messages using S-BERT\footnote{We use \texttt{all-mpnet-base-v2} as the current best performing model for sentence representation.}~\cite{DBLP:journals/corr/abs-1908-10084}. 
If the cosine distance between the vector representations of the messages in a pair is below a threshold, we consider the messages to be semantically diverse. 
We empirically found that a threshold value of 0.5 is sufficient to study dialogue diversity in \ArgSciChat.
Qualitatively similar results obtained with different thresholds are reported in Appendix~\ref{appendix:topical_diversity}. 

We report the average percentage of semantically diverse pairs in each group over all papers with multiple dialogues.  
About 90\% of messages in group G1, 63\% in group G2, and 43\% in group G3 are semantically diverse. 
These results suggest that our dialogue formulation yields diverse dialogues even on identical papers. 
This property of \ArgSciChat is important when its dialogues are used to evaluate dialogue agents \cite{reddy2019coqa}.  

\paragraph{Supportive Facts}
Up to two supportive facts are used as groundings for \e's messages. 
\begin{table}[!t]
\small
\centering
\begin{tabular}{lc}
    \toprule
    \textbf{Property}                 & \textbf{Value}     \\ 
    \midrule
    1-Fact messages             & 61.1 \%       \\
    2-Fact messages             & 38.9  \%    \\
    Avg. sentence distance between facts         & 5.8       \\
    \midrule
     Facts from abstract      & 38.6  \%  \\
    Facts from introduction  & 61.4   \% \\
    \bottomrule
\end{tabular}
\caption{Studied statistics about supportive facts in \ArgSciChat.}
\label{tab:fact_analysis}
\end{table}
Table~\ref{tab:fact_analysis} shows that about 61.1\% (129 out of 211) of these messages are grounded in one supportive fact and 38.9\% (82 out of 211) on two supportive facts. 
For the two-facts messages, the average sentence-based distance between the facts in the scientific paper is 5.8. 
This observation indicates that the supportive facts in the two-facts messages are non-adjacent sentences in the paper.

We also observe that a large portion of all the supportive facts is from the introduction section. 
In particular, 61\% of supportive facts are from the introduction section, while 39\% are from the abstract. 
This shows that both abstract and introduction sections of the scientific paper are explored in the dialogues.

\begin{figure}[!t]
    \centering
    \includegraphics[width=0.50\textwidth, trim=0 0 0 38]{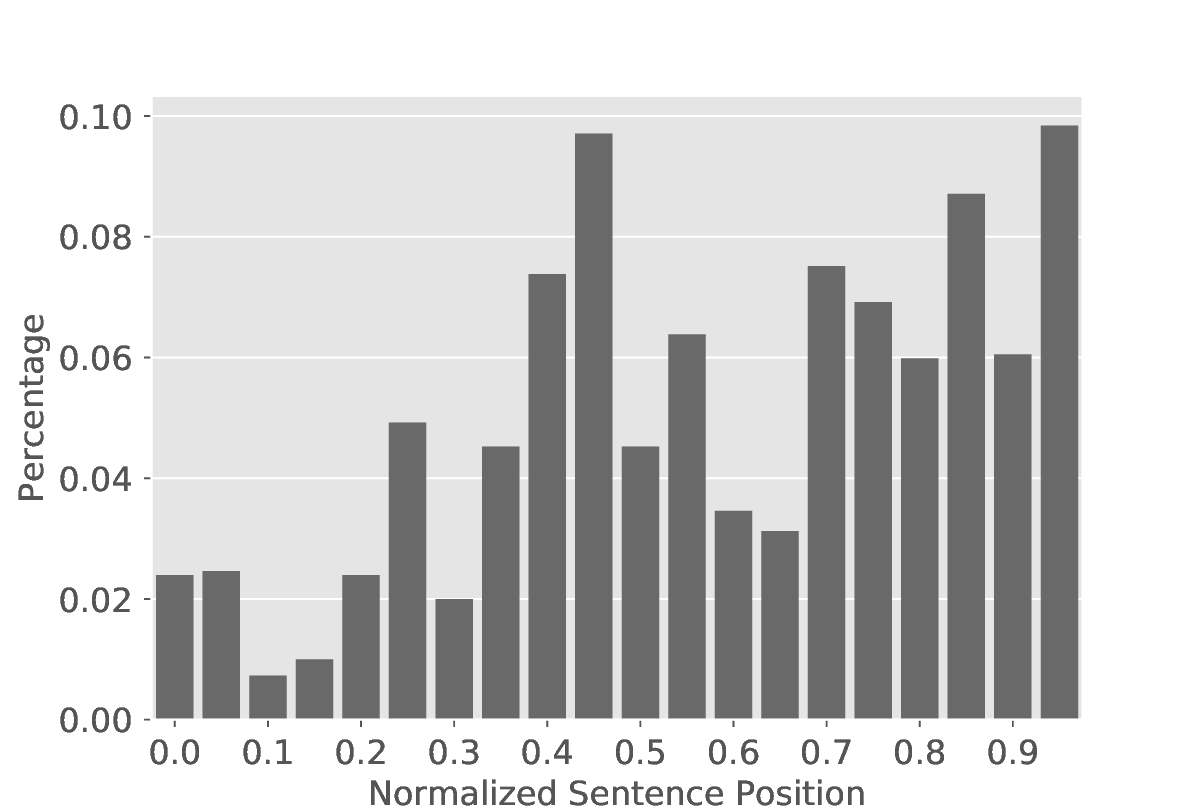}
    \caption{The distribution of supportive facts over sentences of the introduction section.} 
    \label{fig:facts_clusters}
\end{figure}

Finally, \e selects most of the supportive facts from the introduction section, denoting the exploratory intent of \p.
We study the distribution of supportive facts over sentences in the introduction section.
We view the introduction section at the sentence level. We compute the supportive fact distribution by normalizing each sentence position concerning the total number of sentences in the introduction section. Figure~\ref{fig:facts_clusters} shows the results.
We observe that supportive facts mostly are from the middle (35\%-55\%) and last (70\%-100\%) sentences of the introduction section. The middle part of the introduction section describes the research method used in the corresponding scientific paper. The last part reports experiments and findings. We conclude that dialogues explore the scientific contributions of a scientific paper from different scientific views.

\paragraph{EXP and ARG Intents}
\label{paragraph:role-specific_behaviours}


To study to what degree the dialogues are exploratory and argumentative, we instruct four NLP experts to annotate all the sentences from \ArgSciChat dialogues. Specifically, each expert annotates 25\% of \ArgSciChat sentences. We remark that these experts are not from those that participated in our dialogue collection study. As instruction, we give them our intent definitions and their example sentences (Table~\ref{tab:action_set}). For the sake of the annotation budget, we compute the inter-annotator agreement on 289 sentences of 10 randomly selected dialogues.
The average Fleiss' Kappa \cite{fleiss} is 0.89, and the Krippendorf's alpha \cite{krippendorff2011computing} is 0.83, showing that the quality of annotations is high enough to be used in our analysis of the EXP and ARG intents of dialogues in \ArgSciChat. 

Table \ref{table:action_distribution} reports the frequency of each intent label on all the sentences expressed by \p and \e from all dialogues in \ArgSciChat.  
\p is more inclined to ask questions (AI), whereas \e reports about facts from paper content (GI). 
This is compatible with our definitions for these roles, showing that our dialogue formulation was comprehensive for the expert subjects to follow.  
\p expresses an opinion (GO) more often than \e by 10\%. 
Both \p and \e ask for opinion (AO) with a similar frequency ($\approx$6\%).  
\e switches the topic of conversation (ST) more frequently than \p by about 6\%, which is because \e has access to more content of the scientific papers than \p. 
Moreover, according to our formulation, \e should drive the conversation to prevent any dead end. 
By grouping intent labels according to their corresponding intent categories (the EXP and ARG columns in Table~\ref{table:action_distribution}), we observe that ARG intents represent nearly 23\% of all the sentences. In comparison, EXP intents represent about 77\% of messages. 
These results support the correctness of our dialogue collection methodology in achieving the goal of collecting dialogues that demonstrate EXP and ARG dialogues in a scientific domain. 
\begin{table}[!t]
\small
\centering
\begin{tabular}{lccccc|cc}
\toprule
 \textbf{Role}   & \textbf{AI} & \textbf{ST} & \textbf{GI} & \textbf{GO} & \textbf{AO} & \textbf{EXP} & \textbf{ARG} \\ 
\midrule
\p   & 63.3    & 1.2        & 7.6        & 21.9        & 6.1         & 72.0        & 28.0         \\
\e  & 1.0         & 7.6        & 72.7        & 11.8        & 6.9         & 81.3        & 18.7         \\ \midrule
Total & 29.5        & 4.7        & 42.9        & 16.4        & 6.5         & 77.1        & 22.9         \\ 
\bottomrule
\end{tabular}
\caption{The frequency (\%) of intents from each dialogue role. ``Total'' is the total frequency of each intent. 
}
\label{table:action_distribution}
\end{table}

\subsection{Using ArgSciChat for Evaluating Dialogue Agents} 
\label{sec:automatic_reply_generation}

\begin{table}[!t]
\small
\centering
    \begin{tabular}{@{}lcccc@{}}
    \toprule
    \textbf{Model}  & \textbf{Fact-F1} & \textbf{Message-F1} & \textbf{BScore} & \textbf{Mover} \\ 
    \midrule
    TF-IDF  & 16.22 &  - & - & - \\ 
    S-BERT  & 13.65 & - & - & - \\
    \midrule
    LED(Q,P)  & 10.58 & 14.25 & 85.85 & 2.25\\
    LED(Q,P,H) & 8.50 & 16.14 & 86.00 & 4.54 \\
    LED(Q,F)  & - & 19.54 & 86.64 &  8.53\\ 
    \midrule
    Human & 51.26 & - & - & -\\
    \bottomrule
    \end{tabular}
\caption{LED's scores for fact selection (Fact-F1) and response generation (Message-F1, BScore, and Mover). }
\label{tab:led_results}
\end{table}

\begin{table*}[!h]
    \centering
    \begin{tabular}{ll}
    \toprule
    \textbf{Query}: & \emph{Sounds promising! So how well do the models perform?} \\
    \midrule
         \textbf{LED(Q,P)}: & \emph{We use a neural network to train the neural network to generate neural networks.} \\ 
         \textbf{LED(Q,P,H)}: & \emph{We use a generic, KB-agnostic annotation approach.} \\ 
         \textbf{LED(Q,F)}: & \emph{They use their models to improve annotation speed. What do you think about this?} \\
         \midrule
         \textbf{Reference \e}: & \begin{tabular}[c]{@{}l@{}} \emph{Using these in a user study, we measured an annotation time improvement of up 35\%.}\\
         \emph{Do you think this is useful?}
         \end{tabular}  \\
         \midrule
         \textbf{Reference Facts}: & \begin{tabular}[c]{@{}l@{}} \emph{We show that statistical machine learning models can be used in an}\\\emph{interactive entity linking setting to improve annotation speed by over 35\%.}\end{tabular} \\
    \bottomrule
    \end{tabular}
    \caption{A case study of the response generated by the LED agent when trained on the \ArgSciChat dataset. }
    \label{tab:case-study}
\end{table*}

Table~\ref{tab:led_results} shows the results for both tasks. 
For fact selection, we observe that the performance of LED in terms of Fact-F1 is comparable with the baseline fact selection models and far from human performance on this task. 
This confirms the difficulty of \ArgSciChat for the fact selection task. 
However, one reason for such a low score of LED is the small number of data samples, motivating further research in developing few-shot learning methods for training conversational agents in expert domains.

For response generation, in terms of the Message-F1 metric, LED(Q, P) achieves a 14.25\% token-level F1 score, followed by the LED(Q,P,H) with a 16.14\% score. 
LED(Q,P,H) outperforms LED(Q,P), suggesting that prior exchanged messages in a dialogue can be beneficial in generating a response to an input query. This also shows that dialogues in our dataset are beyond a sequence of unrelated dialogue turns.
LED(Q,F) outperforms LED(Q,P) and LED(Q,P,H) by at least three token-level F1 points.  
This score of LED(Q,F) confirms the importance of supportive facts for generating a high-quality response. 

As a case study, we analyze responses LED generates for a query Q given different inputs.  
Table~\ref{tab:case-study} shows the query, the responses generated by the examined configurations of LED, the response generated by \e, and the supportive facts selected by \e in a dialogue. 
LED(Q,F) identifies Q's intent, which aims to ask for information about ``the speed improvement of the annotation study''. The model generates a response with both EXP and ARG intents.  
In particular, the first sentence in the LED(Q,F)'s response has EXP intent as it provides information to answer the question. The second sentence involves ARG intents as it asks for \p's opinion, heading towards an argumentative dialogue. 
\section{Discussion}
Our analysis of dialogues in \ArgSciChat suggests that our methodology is an effective practice for collecting dialogues in expert domains. 
In this work, we applied our methodology to the NLP scientific domain. 
To demonstrate that the methodology is sound, we have run a pilot study and invited 31 NLP experts. Among the invited experts, 23 of them accepted our invitation and participated in our dialogue collection study.
In future work, the methodology could be further scaled to online community events, poster sessions in online conferences, where scientists chat about their papers. 

The performance of LED on our dataset highlights different characteristics of \ArgSciChat, such as the importance of supportive facts for response generation, the benefit of including dialogue history, and the challenge of generating messages that have both EXP and ARG intents. In particular, these characteristics motivate further research on building conversational agents in expert domains. \old{Our work motivates and fosters further research, in particular on developing few-shot learning for conversational AI in expert domains. 
Indeed, collecting large-scaled annotated data in expert domains is a challenge and expensive. Thus, the state-of-the-art data hungry dialogue agents fail in this domain.}
Our dataset's size testifies to the challenge of incentivizing scientists' contribution to data collection. Recent large language models have shown a remarkable ability for zero- and few-shot learning~{\cite{BrownNIPS2020}}. We believe that even a small-sized but high-quality dataset could contribute significantly to developing dialogue agents in specialized domains. 

\section{Conclusion}
We propose \ArgSciChat, the first argumentative dialogue dataset in the NLP scientific domain. In our dataset, dialogue sentences are in free-form English texts, written by NLP experts, and annotated with intent labels encoding the EXP and ARG intents of scientific argumentation. To collect this dataset, we defined a new methodology that lets scientists introduce their scientific papers, choose the topic and the time of dialogues based on their preferences. In addition to in-depth analysis of \ArgSciChat, we found that \ArgSciChat is a challenge for a recent dialogue agent~\cite{Beltagy2020LongformerTL} for the tasks of supportive fact selection and response generation. In future work, we aim at designing few-shot learning agents in the scientific domain.

\section*{Ethical Consent}
We ask experts to read and confirm a consent concerning data privacy and informed consent before signing up for our tool. In the form, we explicitly state the aim of the study and the later use of collected data. We provide detailed information to the subjects about the personal data information we require for participation and its temporary usage throughout the study. Subjects can request data deletion at any given step of the study. All subjects who agree to sign-up also consent to participate in the study.

\section*{Limitations}

\ArgSciChat is the result of a pilot study concerning 31 invited NLP experts. In particular, \ArgSciChat contains dialogues about 20 scientific papers regarding a few NLP topics. Thus, dialogues in \ArgSciChat are only a small sample of the set of possible dialogues grounded in scientific papers. In particular, several design choices have been considered in our data collection methodology: (a) the topic of a paper; (b) the common background of invited NLP experts; (c) the available content of a paper during a dialogue; and (d) the dialogue setting (e.g., in our implementation, dialogues had a time limit which restricted the number of interactions between subjects). We chose the NLP domain in our study since we (the authors) have expertise in this domain. This choice also facilitated the definition of a pool of NLP experts to participate in our study through our research network.

Furthermore, dialogues in \ArgSciChat are grounded in scientific papers. In particular, we limit the paper's content to the abstract and introduction sections. This choice reduces subjects' effort to act as \e, while also providing enough information to sustain a dialogue. Thus, we do not collect dialogues between subjects concerning other sections of a paper.

\bibliography{anthology,article}
\bibliographystyle{acl_natbib}

\newpage
\appendix\normalsize

\section{Dialogue Collection Supplementary Details} \label{appendix:collection_details}

We provide details about our implementation of the described data collection methodology below. We define a time deadline for each step to ease subjects' synchronization along the data collection process. 

\paragraph{Sign-up}
We require subjects to provide the following contact information: full name and email address. We also need subjects to provide their Google Scholar profiles regarding paper selection. If no Google Scholar profile is available, subjects can still sign-up by manually submitting hyperlinks to the PDF files of papers about which they have enough expertise. The automatic paper retrieval step reports the top five subjects' most recent papers listed in their Google Scholar profile. Subjects are then asked to select two of them to participate in the study.

\paragraph{Booking as \e}
We consider time slots of one hour in length. Upon sign-up, summary information about selected papers and time slots is provided to human subjects. In particular, subjects can add the selected time slots to their calendar to ensure participation.

\paragraph{Booking as \p}
Once the sign-up deadline has been met, our implementation automatically notifies subjects about the next phase via email. Additionally, a unique private authentication code is assigned and provided to each participant. The authentication code is used to identify subjects during the study while ensuring anonymity. We split each \e time slot into three time slots of 20 minutes duration. Subjects are required to book four distinct time slots, each one regarding a different paper. Upon submission, summary information is provided to subjects regarding their final schedule concerning both roles.

\paragraph{Conducting a Dialogue}
Once the time deadline of the "Booking as \p" step has been met, our implementation automatically notifies subjects about their final schedule. By doing so, subjects have exact information about which time slots they have to join. Our implementation also supports an automatic notification system that ensures subjects join each time slot on time.

\paragraph{Why Synchronous Dialogues?}
Given subjects' required level of expertise and to ensure naturalness during dialogue collection, we opted for synchronous dialogues. Indeed, such a choice introduces tight requirements about subjects' availability and corresponding attributable effort. However, asynchronous dialogues can also have potential drawbacks, such as an extended data collection period. Additionally, in an asynchronous dialogue, dialogue properties like the partial observability of \p can be quickly overcome due to the absence of a time limit. Conversely, short synchronous dialogues encourage both subjects to maximize the exchanged information flow in the same way as a natural goal-oriented conversation.

\begin{figure*}[!t]
     \centering
     \begin{subfigure}[b]{0.99\textwidth}
         \centering
         \includegraphics[width=\textwidth]{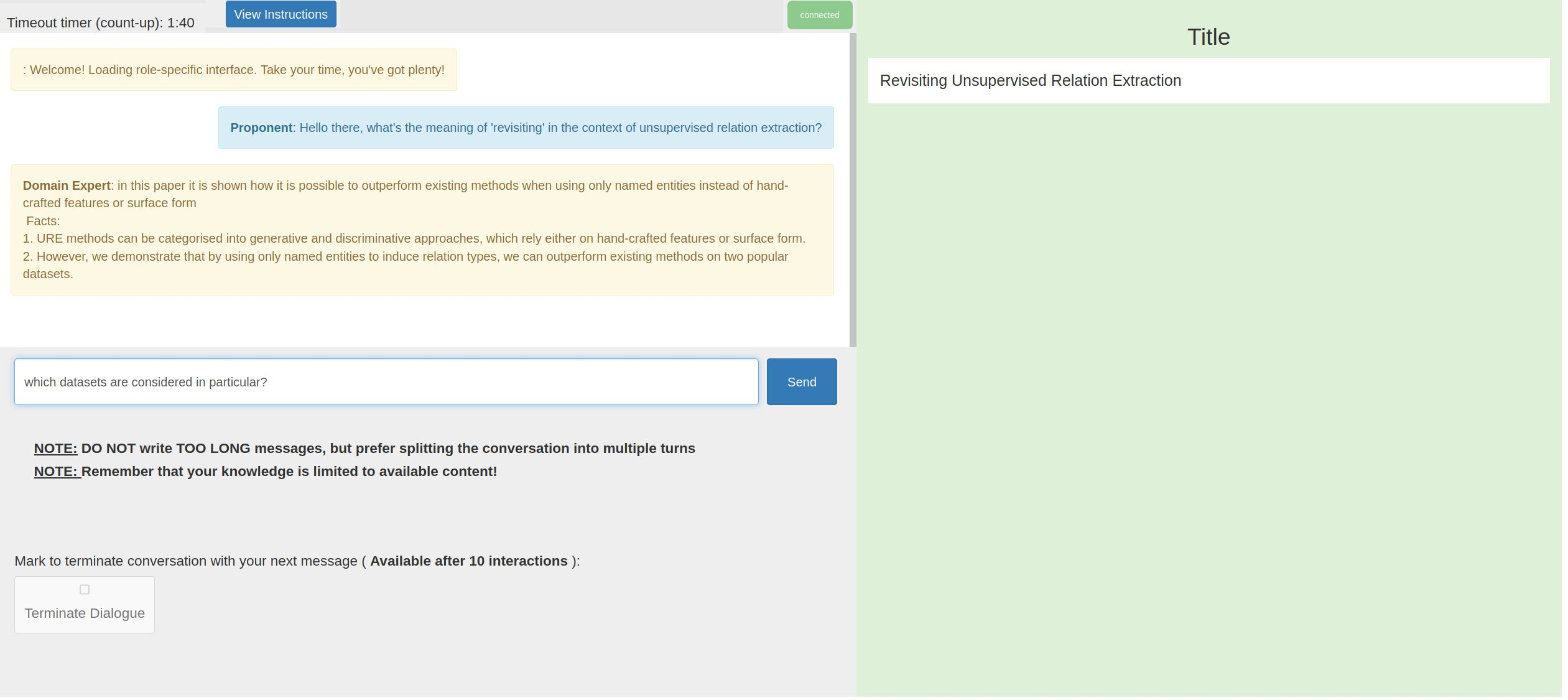}
         \caption{}
         \label{fig:collection_interfaces:proponent}
     \end{subfigure}
     \hfill
     \vspace{0.3cm}
     \begin{subfigure}[b]{0.99\textwidth}
         \centering
         \includegraphics[width=\textwidth]{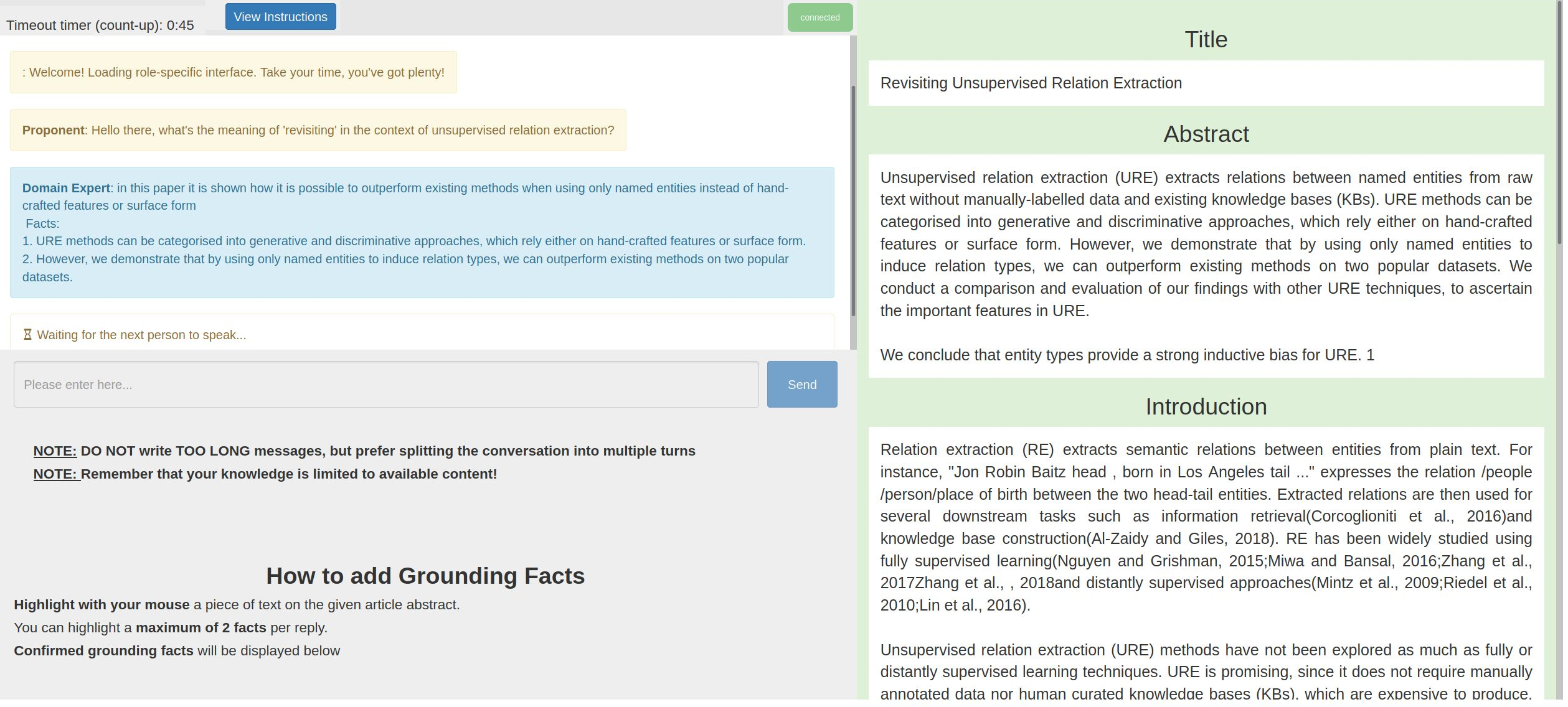}
         \caption{}
         \label{fig:collection_interfaces:domain_expert}
     \end{subfigure}
     \hfill
     
    \caption{Data collection interfaces for (a) \p and (b) \e. On the right side, the title, abstract and introduction sections of the paper are reported for \e. Conversely, \p's view is solely restricted to the paper title.}
    \label{fig:collection_interfaces}
\end{figure*}

\paragraph{Data Collection Interfaces}
We develop role-specific collection interfaces based on the Mephisto library\footnote{\url{https://github.com/facebookresearch/Mephisto}}.
We opt for a standalone data collection application that does not impose additional participation steps apart from initial data submission.
Upon connection of both subjects for a dialogue, the corresponding paper content is loaded, and dialogue roles are attributed to each subject.
Coherently with the given formulation, no worker payment was considered, but an award system is devised to encourage attendance.
To avoid a \e message without supportive facts, the system notifies when no facts have been highlighted upon a message sent. Additionally, a hint system encourages both subjects to suggest discussing previous dialogue turns encouraging opinions exchange. Figure \ref{fig:collection_interfaces} reports the data collection interfaces for both dialogue roles. Both interfaces reflect the asymmetric nature of the dialogue by restricting information to individual roles. 

\section{Dialogue Diversity Supplementary Details} \label{appendix:topical_diversity}

Table~\ref{tab:sentencebert} shows the diversity scores for each group of messages across different diversity thresholds.

\begin{table}[!h]
\small
\centering
\begin{tabular}{ccccccc}
\toprule
\textbf{Group} & & \textbf{0.3}  & \textbf{0.4}  & \textbf{0.5}  & \textbf{0.6}  & \textbf{0.7}  \\ \midrule
G1   & & 0.77 & 0.87 & 0.92 & 0.96 & 0.98 \\
G2   & & 0.43 & 0.48 & 0.63 & 0.73 & 0.68 \\
G3   & & 0.18 & 0.28 & 0.43 & 0.60 & 0.81 \\ \bottomrule
\end{tabular}
\caption{Average topical diversity on dialogues on identical papers.
The numbers in the header of columns represent thresholds values. 
}
\label{tab:sentencebert}
\end{table}

\section{Experiment Supplementary Details} \label{appendix:experiments}

In this section, we report additional details regarding our experimental setup and the described tasks.

\paragraph{Setup}

As in \citet{Dasigi2021ADO}, multiple inputs are concatenated together with a special separation token. In our experimental setup, inputs are concatenated based on the following order: (i) query (Q); (ii) dialogue history (H); (iii) scientific paper (P). We set the LED global attention mask to take into account each individual input as in \cite{Dasigi2021ADO}. 
In contrast to \cite{Dasigi2021ADO}, the lack of LaTeX source files forbids the automatic retrieval of paper section paragraphs. Instead, we work at the sentence level. The average training time is ${\sim}40-60$ seconds per epoch depending on the given input combination. Experiments are conducted on a NVidia GeForce 2080ti 11 GB.

\paragraph{Model Configuration}

Table \ref{table:appendix:experiments:model_configuration} reports the main hyper-parameters of the LED model. As in \cite{Dasigi2021ADO}, we employ the HuggingFace model \texttt{allenai/led-base-16384}. Because we only use the abstract and introduction sections of a scientific paper, we re-scale the \textit{attention window size} hyper-parameter according to our maximum document length. Note that the selected hyper-parameter value is the same percentage of the maximum document length as in \cite{Dasigi2021ADO}.

\begin{table}[!th]
    \centering
    \small
    \begin{tabular}{lc}
    \toprule
    \textbf{Hyperparameter}       & \textbf{Value} \\ \midrule
    Attention dropout     & 0.1   \\
    Attention window size & 700   \\
    Optimizer              & Adam  \\
    Learning rate         & $5 \cdot 10^{-5}$  \\
    Patience            & 20 \\ \bottomrule
    \end{tabular}
    \caption{The LED model hyper-parameters.}
    \label{table:appendix:experiments:model_configuration}
\end{table}

\end{document}